\documentclass[sigconf]{acmart}
\usepackage{balance}
\usepackage[ruled,vlined]{algorithm2e}
\usepackage{multirow}

\AtBeginDocument{%
  \providecommand\BibTeX{{%
    \normalfont B\kern-0.5em{\scshape i\kern-0.25em b}\kern-0.8em\TeX}}}

\copyrightyear{2024}
\acmYear{2024}
\setcopyright{acmlicensed}
\acmConference[MRAC '24] {Proceedings of the 2nd International Workshop on Multimodal and Responsible Affective Computing}{November 1, 2024}{Melbourne, VIC, Australia.}
\acmBooktitle{Proceedings of the 2nd International Workshop on Multimodal and Responsible Affective Computing (MRAC '24), November 1, 2024, Melbourne, VIC, Australia}
\acmISBN{979-8-4007-1203-6/24/11}
\acmDOI{10.1145/3689092.3689411}

\begin{document}

\title{Learning Noise-Robust Joint Representation for Multimodal \\ Emotion Recognition under Incomplete Data Scenarios}


\author{Qi Fan}
\authornote{Equal contributions.}
\affiliation{%
  \institution{Inner Mongolia University}
  \city{Hohhot}
  \country{China}
}
\email{fanqi1203@foxmail.com}
  
\author{Haolin Zuo}
\authornotemark[1]
\affiliation{%
  \institution{Inner Mongolia University}
  \city{Hohhot}
  \country{China}}
\email{zuohaolin\_0613@163.com}

\author{Rui Liu}
\authornote{Corresponding Author.}
\affiliation{%
  \institution{Inner Mongolia University}
  \city{Hohhot}
  \country{China}}
\email{liurui\_imu@163.com}

\author{Zheng Lian}
\affiliation{%
  \institution{Institute of Automation, Chinese Academy of Sciences}
  \city{Beijing}
  \country{China}}
\email{lianzheng2016@ia.ac.cn}

\author{Guanglai Gao}
\affiliation{%
  \institution{Inner Mongolia University}
  \city{Hohhot}
  \country{China}
}
\email{csggl@imu.edu.cn}
\renewcommand{\shortauthors}{Qi Fan, Haolin Zuo, Rui Liu, Zheng Lian \& Guanglai Gao}

\begin{abstract}
Multimodal emotion recognition (MER) in practical scenarios is significantly challenged by the presence of missing or incomplete data across different modalities. To overcome these challenges, researchers have aimed to simulate incomplete conditions during the training phase to enhance the system's overall robustness. Traditional methods have often involved discarding data or substituting data segments with zero vectors to approximate these incompletenesses. However, such approaches neither accurately represent real-world conditions nor adequately address the issue of noisy data availability. For instance, a blurry image cannot be simply replaced with zero vectors, while still retaining information. To tackle this issue and develop a more precise MER system, we introduce a novel noise-robust MER model that effectively learns robust multimodal joint representations from noisy data. This approach includes two pivotal components: firstly, a noise scheduler that adjusts the type and level of noise in the data to emulate various realistic incomplete situations. Secondly, a Variational AutoEncoder (VAE)-based module is employed to reconstruct these robust multimodal joint representations from the noisy inputs. Notably, the introduction of the noise scheduler enables the exploration of an entirely new type of incomplete data condition, which is impossible with existing methods. Extensive experimental evaluations on the benchmark datasets IEMOCAP and CMU-MOSEI demonstrate the effectiveness of the noise scheduler and the excellent performance of our proposed model. Our project is publicly available on \href{https://github.com/WooyoohL/Noise-robust_MER}{https://github.com/WooyoohL/Noise-robust\_MER}.
\end{abstract}



\maketitle

\begin{figure}[h]
\centering
\centerline{\includegraphics[width=1.1\linewidth]{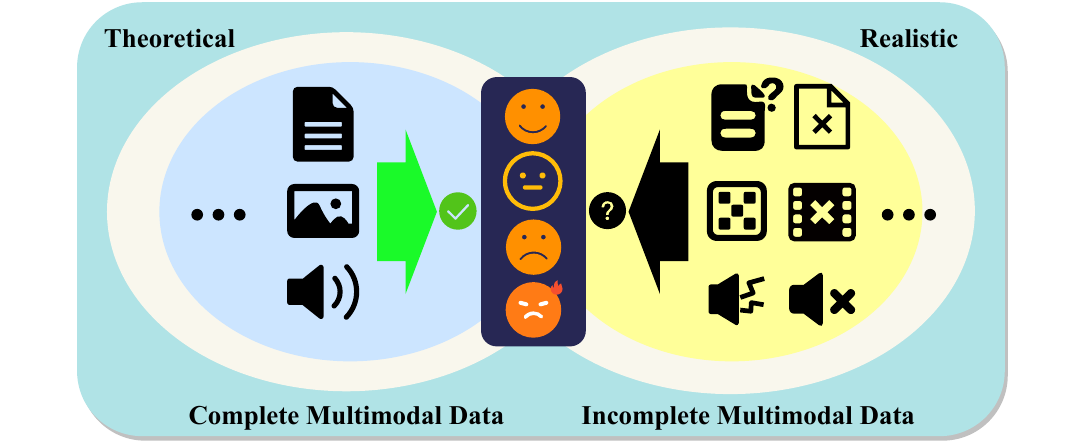}}
\caption{The incompleteness of multimodal data in realistic scenarios poses a significant challenge to multimodal emotion recognition.}
\label{fig: example}
\end{figure}
\section{Introduction}
\label{sec: intro}
Multimodal emotion recognition (MER) aims to take multimodal signals, including text, audio, and visual, as input to predict the emotion category \citep{zuo2023}. As a cutting-edge technology, it is widely used in various scenarios, including virtual intelligent assistants, robot customer service, and other applications, offering tailored and empathetic user experiences by responding appropriately to the emotional cues of users. In the research of MER, remarkable performance depends heavily on the complete multimodal data and robust joint representation learning \citep{hazarika2020}.
However, in realistic scenarios, data often becomes incomplete due to two main cases: either data is \textbf{absolutely absence} caused by sensor malfunction \citep{zhao2021}, or it is \textbf{partially incompleteness} attributed to diminished network bandwidth or various forms of noise interference \citep{liu2022, zhou2021}, etc. We collectively call these data noisy or incomplete data, which present huge challenges for multimodal emotion recognition \citep{zhao2021}.

\begin{figure*}[h]
\centering
\centerline{\includegraphics[width=0.9\linewidth]{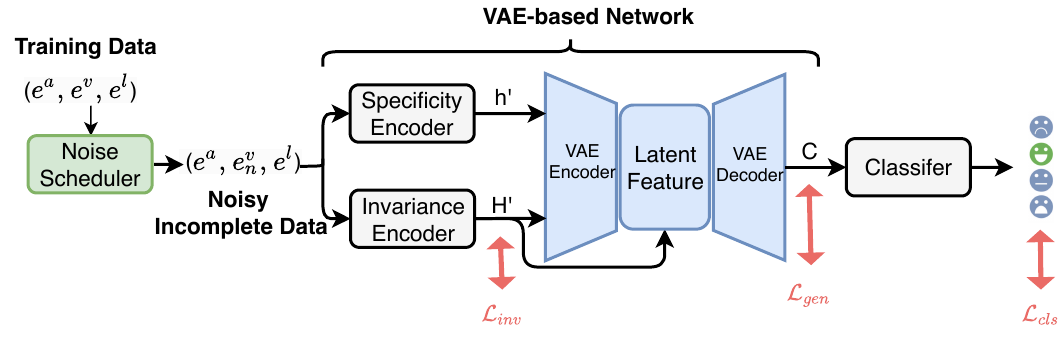}}
\caption{The structure of our NMER model, including noise scheduler, VAE-based network, and the classifier. $\mathcal{L}_{inv}$ stands the invariant loss, $\mathcal{L}_{gen}$ is the generation loss, $\mathcal{L}_{cls}$ refers to the classification loss.}
\label{fig: NMER Model}
\end{figure*}

In recent years, researchers have tried to enhance the robustness of MER by simulating incomplete data during training. 
Previous researchers proposed two methods to simulate incomplete data according to the usual intuition: 1) set feature vectors of missing data to zero: in some situations, zero is a value that signifies ``no information'' or ``lack of information'' \citep{zhao2021, zuo2023}, 2) discard data randomly with a predefined probability: dropped data can be regarded as completely missing \citep{lian2023, chen2020}. 
On this basis, MER work in the face of incomplete data focuses on two main areas: 1) missing data completion \citep{Cai2018, Suo2019, Du2018, lian2023} and 2) multimodal joint representation learning with available data \citep{Han2019, Pham2019}. 
For example, 
Cai et al. \citep{Cai2018} proposed to use adversarial learning to generate missing modality images.
Zeng et al. \citep{zeng2022} proposed an ensemble learning method to use several models to solve the missing problem jointly. 
Zhao et al. \citep{zhao2021} proposed a Missing Modality Imagination Network (MMIN) using AutoEncoder and cycle consistency construction to learn joint representations while predicting missing modalities. 
Zuo et al.\citep{zuo2023} proposed introducing the modality invariant feature into MMIN to learn robust multimodal joint representations. 
Liu et al. \citep{liu2024} combined contrastive learning and invariant features to imagine the feature of the missing modality.
The above work lays a solid foundation for MER work under incomplete data. Note that the framework that combines both \textit{missing data completion} and \textit{multimodal joint representation learning} has become the mainstream scheme in the research of this area.

However, the above approach faces two main issues, including 1) the method of simulating incomplete data is unreasonable and impractical: traditional data setup methods set parts of the feature to zero vectors or discard some data directly at a set percentage. These methods neither reliably simulate real-world scenarios nor reserve the availability of incomplete data. Moreover, assuming the data contains some type or level of noise, is not it better to try to mine useful information from it rather than just discard it? and 2) the structure is redundant: traditional model structure learns a multimodal joint representation in the process of completing the missing data \citep{zuo2023}, and such a process tends to cause the errors generated in the first step to have a negative impact on the second step.  Whether robust multimodal joint feature representations can be learned directly from noisy data?

To answer the above two questions, we propose a novel Noise-robust MER model, termed NMER.
Specifically, we design a noise scheduler at the embedding level. It creates noisy training and testing data by adding various types and intensities of noise to the embedding, thus simulating the influence of diverse incomplete situations in realistic scenarios. It is worth mentioning that we explored an incomplete condition of three new modalities through our noise scheduler, which was impossible to achieve in previous works. Then, we present a Variational AutoEncoder (VAE) \citep{kingma2013, sohn2015} based multimodal joint representation learning network to reconstruct robust multimodal joint representations from the noisy data. 
In this way, we simulate incomplete data in realistic scenarios, make full use of the valuable information of the existing noisy data, and then leverage the powerful generative capabilities of VAE to reconstruct robust multimodal joint representations from noisy data and achieve multimodal emotion recognition. We conduct experiments on the benchmark dataset IEMOCAP \citep{Busso2008} and CMU-MOSEI \citep{liang2021}, which are widely used. 
Our code has been released in the supplementary material.

The main contributions of this work are as follows: 

$\bullet$ We propose a noise-robust MER model, termed NMER, to generate robust multimodal joint representations under noisy incomplete data and proceed with multimodal emotion recognition. 

$\bullet$ We explore a new method to simulate realistic data under various incomplete conditions utilizing a noise scheduler while enabling a new condition with the incompletion of all three modalities.

$\bullet$ Experimental results under various noise types and intensity conditions show that our NMER outperforms most baselines and demonstrates robustness in the face of incomplete data.

\section{NMER: Methodology}
\label{sec: model}

\subsection{Overall Architecture}
The overall architecture of NMER is illustrated in Figure~\ref{fig: NMER Model}, which consists of 1) \textit{Noise Scheduler}; 2) \textit{VAE-based Network}; and 3) \textit{Classifier}.
Specifically, we first employ our \textit{Noise Scheduler} to add configurable noise to original embeddings to get incomplete data and send them into the VAE-based network. Then, the \textit{VAE-based Network} seeks to extract the useful features to reconstruct the robust multimodal joint representation.
At last, the joint representation will be fed into \textit{Classifier} to predict the final emotional result.

\subsection{Noise Scheduler}
\subsubsection{Rationale and Noise Selection}\

Inspired by the previous work \citep{ho2020}, we can obtain a blurred picture by gradually adding Gaussian noise to the picture. Similarly, we can obtain a "noisy embedding" by adding noise to the embedding \citep{sietsma1991creating, miyato2017adversarial}, which enhances the robustness of the model while simulating the modality incomplete condition.
We examine two common noise types as examples: Gaussian noise, as described by C. F. Gauss in \citep{gauss1877}, and impulse noise, following the research of Kim et al. \citep{kim1995}

Gaussian noise is extensively employed to simulate errors and disturbances occurring in natural and technical processes due to its distinct statistical characteristics, namely a well-defined mean $\mu$ and standard deviation $\sigma$ that conform to a normal distribution. The selection of this type of noise is grounded in a crucial observation: deviations influenced by various factors tend to approximate a normal distribution in most natural and technological systems. Consequently, the use of Gaussian noise in practical applications not only maintains statistical rigor but also ensures that models exhibit greater robustness and adaptability when confronted with the complexities of real-world data \citep{bishop2006, hastie2009}.

Impulse noise is another typical noise form that often appears in signal processing. Due to its ability to simulate brief and intense disturbances, it is widely used to model sudden anomalies in data transmission or sensor inputs. Additionally, impulse noise can also be employed to enhance the capability of the model to recognize and process extreme data points \citep{mafi2019, liu2019impulsive}.

The integration of noise at the embedding level is a pivotal aspect of simulating realistic scenarios. Previous works have prove its effectiveness and robustness facing real situtaions. For the incomplete modality problem, we can choose various techniques for corrupting original data, such as text (masking, reversing word order), audio (adding reverberation or noise), and video (introducing frame loss or blur). However, quantifying the specific impact of such corruption on each modality is quite difficult. For instance, determining the equivalent effect of adding a certain decibel level of noise to audio, or masking corresponding words, if a frame in a video is completely missing, is particularly difficult. Therefore, we introduce noise on the embedding to strike a balance between realism and analytical feasibility. 

Based on the reasons above, we select those two types of noise on the embedding level, bring quantitative loss of information through control of the noise level, and simulate the negative impact on information from noise in reality.
To be clear, our noise scheduler is designed to separate the processes of generating and adding noise. Users have the freedom to select or customize any type of noise, provided it can be created as a tensor using a programming language. This design ensures that the choice of noise type does not impact the addition process, offering users significant flexibility to adapt to various data processing needs.

\subsubsection{Noisy Embedding Construction}\

Following the diffusion process in \citep{ho2020}, we construct noisy embedding through a similar process.
Given complete multimodal data samples, encompassing acoustic, visual, and lexical components (denoted as $a$, $v$, and $l$ respectively), the embeddings are noted as \(\boldsymbol{E^a}\), \(\boldsymbol{E^v}\), and \(\boldsymbol{E^l}\). We use \(\boldsymbol{E^a}\) as an example to illustrate the process of making incomplete data.

For constructing the noisy embedding using Gaussian noise, we build a noise schedule across time steps represented by \(\mathrm{\beta_{s}}\):

\begin{equation}
    \mathrm{\beta_{s}} = (\beta_1, \dots, \beta_T)
\end{equation}
It initializes a sequence of values starting from a small positive number \(\beta_{\text{start}}\) (denoting a low noise level) and progressively increasing towards a larger number \(\beta_{\text{end}}\) (denoting a high noise level) after \(T\) steps. The total number of steps \(T\) influences the gradual transition of data toward a noise-dominated state, where \(T\) should be an integer greater than zero. The \(\beta_{\text{start}}\) and \(\beta_{\text{end}}\) are fixed to 0.001 and 0.1 generally refer to previous work \citep{ho2020}. The embedding is noted as \(\boldsymbol{E^a_0}\) at time step 0. We then use Algorithm~\ref{algorithm1} to generate the noisy embedding \(\boldsymbol{E^a_T}\).

\begin{algorithm}[]
\SetAlgoLined
\textbf{Input:} Original embedding \(\boldsymbol{E^a_0}\), number of steps \(T\), noise schedule parameters \(\beta_1, \beta_2, \ldots, \beta_T\)\\

\For{$t = 1$ \KwTo $T$}{
 Sample noise vector $\boldsymbol{\epsilon_t} \sim \mathcal{N}(0, 1)$\;
 Update $\boldsymbol{E^a_T} \gets \sqrt{1-\beta_t} \boldsymbol{E^a_{t-1}} + \sqrt{\beta_t} \boldsymbol{\epsilon_t}$\;
}
\textbf{Output:} Noisy embedding \(\boldsymbol{E^a_T}\)\\
\caption{Process for Generating Noisy Embedding \(\boldsymbol{E^a_T}\)}
\label{algorithm1}
\end{algorithm}

However, this method is computationally intensive. To optimize this process for directly obtaining \(\boldsymbol{E^a_T}\) from \(\boldsymbol{E^a_0}\) without iterating through each intermediate step from \(\boldsymbol{E_1}\) to \(\boldsymbol{E_{T-1}}\), we precompute the cumulative product of  \(1 - \beta\) terms, denoted as \(\bar{\alpha}_t\):

\begin{equation}
\bar{\alpha}_T = \prod_{t=1}^{T} \left( 1 - \beta_t \right)
\label{alpha-}
\end{equation}
It significantly reduces computational complexity, which reflects the total variance retained in the data up to step \( T \). Utilizing \(\bar{\alpha}_t\), the noisy embedding \(\boldsymbol{E^a_T}\) is obtained from \(\boldsymbol{E^a_0}\) with a single update:

\begin{equation}
\boldsymbol{E^a_T} = \sqrt{\bar{\alpha}_T} \boldsymbol{E^a_0} + \sqrt{1 - \bar{\alpha}_T} \boldsymbol{\epsilon}
\label{cal noise}
\end{equation}
where \(\boldsymbol{\epsilon}\) is a freshly sampled noise vector from a Gaussian distribution, \(\boldsymbol{\epsilon} \sim \mathcal{N}(0, 1)\). This method enables an efficient bypass of the sequential update steps, directly synthesizing the noisy vector at any time step \( T \) that we desired.

For impulse noise, the sampled noise vector \(\boldsymbol{\epsilon}\) consists only of random values 1s and -1s. An appearance frequency \(p\) is employed to adjust its values, setting \(1 - p\) percent of them to zero and leaving the remaining \(p\) percent unchanged. This adjustment introduces noise to \(p\) percent of data. Then we still use Equation~\ref{cal noise} to get \(\boldsymbol{E^a_T}\).

\subsubsection{Noise Intensity Calculation}\

In the noisy embedding construction process, the $\beta_{\text{start}}$, $\beta_{\text{end}}$, and $T$ control the intensity of noise jointly. Therefore, we better quantify the noise intensity by calculating the signal-to-noise ratio (\textit{SNR}) between the original and noisy data, in other words, the intensity ratio between noise and raw data.

In Equation~\ref{cal noise}, we regard $\sqrt{\bar \alpha_T}$ and  $\sqrt{1- \bar \alpha_T}$ as the power of the original data part and the noise part, separately.
According to Equation~\ref{alpha-} and~\ref{snr}, it's easy to know that when $\sigma^2$ is invariant, the \textit{SNR} value shares the opposite trend with $T$ growing. Hence we can obtain a higher level of noise by increasing the value of $T$.

\begin{equation}
    \mathrm{SNR} = \frac{P_{original}}{P_{noise}} = \frac{\bar \alpha_T \sigma^2}{1- \bar \alpha_T}
\end{equation}
where the $\sigma_2$ is the variance of the original data. More intuitively, \textit{SNR} is always represented in decibels(\textit{dB}):
\begin{equation}
    \mathrm{SNR}_{dB} = 10 \log_{10} \left( \frac{P_{original}}{P_{noise}} \right) = 10 \log_{10} \left( \frac{\bar \alpha_T \sigma^2}{1- \bar \alpha_T} \right)
    \label{snr}
\end{equation}
According to Equation~\ref{alpha-} and~\ref{snr}, it's easy to know that when $\sigma^2$ is invariant, the \textit{SNR} value shares the opposite trend with $T$ growing. Hence we can obtain a higher level of noise by increasing the value of $T$.
Due to the variance differences across different datasets and modalities, we select a distinct T-value for each modality to maintain a uniform noise level. For instance, within the CMU-MOSEI dataset, we adjust the $T$-value for $a$, $v$, and $l$ modalities to [140, 60, 5] respectively to introduce noise at -10\textit{dB}.
By doing so, we are able to impose a similar level of noise on different datasets and modalities, which simulate diverse levels of information loss.

\subsection{VAE-based Network}
The VAE-based Network includes Specificity and Invariance Encoders \citep{zuo2023} and the VAE Module. The Specificity Encoders, employing LSTM \citep{LSTM} and TextCNN \citep{TextCNN} structures, are tasked with extracting modality-specific emotion features \(\boldsymbol{h'}\) from each modality, which is achieved by mapping the embeddings of different modalities into distinct vector spaces. Concurrently, the Invariance Encoder, utilizing a linear structure, extracts modality-invariant emotion features \(\boldsymbol{H'}\) across various modalities by mapping these embeddings into a unified vector space \citep{zuo2023,liu2024}. 

Upon sending the concatenated features \(\boldsymbol{h'}\) and \(\boldsymbol{H'}\) into the VAE model, the VAE Encoder compresses and maps them, yielding mean and variance parameters within the latent space. This step effectively translocates the feature representation from the original data space to a probabilistic distribution in the latent space. A stochastic sampling process, facilitated by the reparameterization trick \citep{kingma2013}, then generates samples that reflect this latent space distribution. These samples embody the latent representations of the input features.

Subsequently, these latent variables pass through the decoder network. The decoder reconstructs the feature, remapping it to the original data space, thereby producing a reconstructed multimodal joint representation denoted as \(\boldsymbol{C}\). Notably, the invariant feature \(\boldsymbol{H'}\) plays a pivotal role during decoding, guiding the model to focus on the common emotional features in the multimodal data.

Noisy features are transformed into normal ones throughout the compression, sampling, and reconstruction phases, which provide the denoising effect. The final output, the multimodal joint representation \(\boldsymbol{C}\), is then fed into the Classifier to derive the result.

\subsection{Loss Functions}
As shown in Fig. \ref{fig: NMER Model}, the total loss $\mathcal{L}$ for NMER includes three parts: $\mathcal{L}=\lambda_1\mathcal{L}_{\text{gen}} + \lambda_2\mathcal{L}_\text{inv} + \lambda_3\mathcal{L}_\text{cls}$, where $\lambda_s$ are the balance factors.

Generation loss \(\mathcal{L}_{\text{gen}}\) aims to calculate the distance between the generation result \(\boldsymbol{C}\) (from incomplete data) and the multimodal joint representation \(\boldsymbol{\hat{C}}\) (from complete data).

Note that the $\mathcal{L}_{\text{gen}}$ consists of two items, that are $\mathcal{L}_\text{kl}$ in Equation~\ref{KLloss} and $\mathcal{L}_\text{mse}$ in Equation~\ref{MSEloss}.
$\mathcal{L}_\text{kl}$ aims to make the hidden variables generated by the encoder conform to the standard normal distribution, while $\mathcal{L}_\text{mse}$ seeks to make the generated multi-modal joint representation more similar to the target that extracted from complete data, 
\begin{equation}
\mathcal{L}_\text{kl}=\!-\frac{1}{2}\left(\log \sigma^2\!\!-\!\sigma^2\!\!-\!\!\mu^2\!+\!1\right)
\label{KLloss}
\end{equation}
\begin{equation}
\mathcal{L}_{\text{mse}} = \frac{1}{N} \sum_{i=1}^{N}(y_i - \hat{y_i})^2
\label{MSEloss}
\end{equation}
where $\sigma$ is the variance of the distribution of the latent vector while the $\mu$ is the mean; $N$ is the total number of the real values, $y_i$ is the real value, and $\hat{y_i}$ is the predicted value. 

The invariant loss \(\mathcal{L}_\text{inv}\) shares the same spirit as in \citep{zuo2023}. It adopts the \textit{MSE} loss style to reduce the distance between the modality-invariant feature \(\boldsymbol{H'}\) during training (under incomplete conditions) and the real modality-invariant feature \(\boldsymbol{H}\). \(\mathcal{L}_\text{cls}\) adopts the Cross-Entropy loss function to measure and minimize the disparities between the predicted and actual emotion category labels.

\section{Experiments and Results}
\label{sec:exp}
\subsection{Data Setup}
We perform experiments on the Interactive Emotional Dyadic Motion Capture (IEMOCAP) \citep{Busso2008} and CMU Multimodal Opinion Sentiment and Emotion Intensity (CMU-MOSEI) dataset \citep{zadeh2018multi}, which are both widely used in the research of MER.

\begin{table}[ht]
\centering
\renewcommand{\arraystretch}{1.4} 
\caption{Emotion distribution and division of two datasets.}
\begin{tabular}{l@{\hspace{7pt}}l@{\hspace{4pt}}c@{\hspace{4pt}}c@{\hspace{4pt}}c@{\hspace{4pt}}}
\toprule
Dataset    & Label & Samples & Train Set & Test Set \\ \midrule
\multirow{4}{*}{IEMOCAP} 
           & Happy         & 1636    & \multirow{4}{*}{4446} & \multirow{4}{*}{3342 (557 samples)} \\
           & Angry         & 1103    &                       &                                        \\
           & Sad           & 1084    &                       &                                        \\
           & Neutral       & 1708    &                       &                                        \\ \midrule
\multirow{2}{*}{CMU-MOSEI} 
           & Positive      & 14842   & \multirow{2}{*}{16265} & \multirow{2}{*}{4643} \\
           & Negative      & 6066    &                        &                      \\
\bottomrule
\end{tabular}
\label{label distribution}
\end{table}

\setlength{\tabcolsep}{0.008 \linewidth} 
\begin{table*}[] \footnotesize \centering
\caption{The results of the comparative studies on the IEMOCAP and CMU-MOSEI dataset, which employs two types of noise. The note ``-10\textit{dB}(Avg)'' means the average performance across six common incomplete conditions under the noise intensity -10\textit{dB}. ``WA'' stands for the weighted accuracy and ``UA'' refers to the unweighted accuracy. Bold values imply the best accuracy on that dataset.}
\renewcommand{\arraystretch}{1.9}
\begin{tabular}{l@{\hspace{10pt}}l@{\hspace{10pt}}cc@{\hspace{5pt}}cc@{\hspace{10pt}}cc@{\hspace{5pt}}cc@{\hspace{10pt}}cc@{\hspace{5pt}}cc@{\hspace{10pt}}cc@{\hspace{5pt}}cc}
\toprule
\multirow{3}{*}{Dataset}   & \multirow{3}{*}{System} & \multicolumn{4}{c}{-10\textit{dB} (Avg)}                                          & \multicolumn{4}{c}{-20\textit{dB} (Avg)}                                          & \multicolumn{4}{c}{-30\textit{dB} (Avg)}                                          & \multicolumn{4}{c}{-40\textit{dB} (Avg)}                                          \\
                           &                         & \multicolumn{2}{l}{Gaussian noise} & \multicolumn{2}{l}{Impulse noise} & \multicolumn{2}{l}{Gaussian noise} & \multicolumn{2}{l}{Impulse noise} & \multicolumn{2}{l}{Gaussian noise} & \multicolumn{2}{l}{Impulse noise} & \multicolumn{2}{l}{Gaussian noise} & \multicolumn{2}{l}{Impulse noise} \\ 
                           &                         & WA               & UA              & WA              & UA              & WA               & UA              & WA              & UA              & WA               & UA              & WA              & UA              & WA               & UA              & WA              & UA              \\ \midrule 
\multirow{7}{*}{IEMOCAP}   & MEN                     & 0.7120           & 0.7175          & 0.7448          & 0.7453          & 0.6747           & 0.6714          & 0.7026          & 0.7133          & 0.6370           & 0.6349          & 0.6758          & 0.6850          & 0.6047           & 0.6004          & 0.6348          & 0.6364          \\
                           & MCTN                    & 0.7215           & 0.7431          & 0.7400          & 0.7531          & 0.6989           & 0.7003          & 0.7270          & 0.7314          & 0.6778           & 0.6712          & 0.7112          & 0.7157          & 0.6674           & 0.6790          & 0.6978          & 0.7041          \\
                           & MMIN                    & 0.7551           & 0.7640          & 0.7750          & 0.7870          & 0.7271           & 0.7394          & 0.7521          & 0.7671          & 0.7155           & 0.7209          & 0.7397          & 0.7516          & 0.7003           & 0.7067          & 0.7226          & 0.7352          \\
                           & IF-MMIN                 & 0.7543           & 0.7655          & 0.7738          & 0.7858          & \textbf{0.7345}  & \textbf{0.7493} & 0.7589          & 0.7719          & 0.7184           & 0.7225          & \textbf{0.7382} & 0.7509          & 0.7048           & 0.7155          & 0.7343          & 0.7452          \\
                           & \textbf{Ours}           & \textbf{0.7598}  & \textbf{0.7675} & \textbf{0.7773} & \textbf{0.7872} & 0.7307           & 0.7406          & \textbf{0.7651} & \textbf{0.7790} & \textbf{0.7197}  & \textbf{0.7284} & 0.7304          & \textbf{0.7532} & \textbf{0.7085}  & \textbf{0.7176} & \textbf{0.7391} & \textbf{0.7503} \\
                           & w/o VAE                 & 0.7382           & 0.7275          & 0.7509          & 0.7631          & 0.7004           & 0.7132          & 0.7467          & 0.7500          & 0.6850           & 0.6882          & 0.7136          & 0.7280          & 0.6793           & 0.6801          & 0.7133          & 0.7238          \\
                           & w/o $\mathcal{L}_{\text{inv}}$                & 0.7401           & 0.7338          & 0.7630          & 0.7712          & 0.7121           & 0.7293          & 0.7489          & 0.7549          & 0.7010           & 0.6942          & 0.7202          & 0.7276          & 0.6843           & 0.6805          & 0.7158          & 0.7208          \\ \midrule
\multirow{7}{*}{CMU-MOSEI} & MEN                     & 0.7285           & 0.6593          & 0.7341          & 0.6638          & 0.6966           & 0.6380          & 0.7010          & 0.6495          & 0.6789           & 0.6099          & 0.6539          & 0.6260          & 0.6528           & 0.5472          & 0.6383          & 0.6135          \\
                           & MCTN                    & 0.7330           & 0.6656          & 0.7455          & 0.6827          & 0.7254           & 0.6451          & 0.7309          & 0.6678          & 0.7231           & 0.6228          & 0.7180          & 0.6454          & 0.7103           & 0.6107          & 0.7007          & 0.6280          \\
                           & MMIN                    & 0.7438           & \textbf{0.6810} & 0.7713          & 0.7035          & 0.7440           & \textbf{0.6676} & 0.7580          & 0.6802          & 0.7414           & 0.6545          & 0.7331          & 0.6540          & 0.7428           & 0.6464          & 0.7277          & 0.6391          \\
                           & IF-MMIN                 & 0.7566           & 0.6750          & 0.7701          & 0.7026          & 0.7431           & 0.6609          & 0.7522          & 0.6703          & 0.7389           & 0.6501          & 0.735           & 0.6558          & 0.7432           & 0.6502          & 0.7393          & 0.6498          \\
                           & \textbf{Ours}           & \textbf{0.7596}  & 0.6760          & \textbf{0.7744} & \textbf{0.7105} & \textbf{0.7454}  & 0.6617          & \textbf{0.7588} & \textbf{0.6831} & \textbf{0.7487}  & \textbf{0.6549} & \textbf{0.7361} & \textbf{0.6572} & \textbf{0.7482}  & \textbf{0.6543} & \textbf{0.7380} & \textbf{0.6530} \\
                           & w/o VAE                 & 0.7439           & 0.6610          & 0.7621          & 0.6800          & 0.7328           & 0.6305          & 0.7483          & 0.6622          & 0.7289           & 0.6300          & 0.7151          & 0.6445          & 0.7018           & 0.6204          & 0.6971          & 0.6210          \\
                           & w/o $\mathcal{L}_{\text{inv}}$                & 0.7521           & 0.6732          & 0.7709          & 0.7034          & 0.7387           & 0.6713          & 0.7511          & 0.6732          & 0.7261           & 0.6659          & 0.7272          & 0.6419          & 0.7230           & 0.6336          & 0.7045          & 0.6223         \\ \bottomrule
\end{tabular}
\label{comp_1}
\end{table*}

In the IEMOCAP dataset, training and testing conditions are consistently following the research 
 \citep{zhao2021}.
For the training set, each sample is subjected to a randomly selected incomplete condition. In contrast, the test set consists of 557 unique samples. Because of the small number of test samples, each sample will be evaluated under six predefined incomplete conditions, resulting in a total of 3342 test samples. In the CMU-MOSEI dataset, both train and test samples are randomly influenced by one incomplete condition, the same as the train set in the IEMOCAP dataset. 

The only difference in data with previous works is that incomplete samples are created with our noise scheduler. 
Besides, the training and testing of the incomplete condition \((\boldsymbol{E^a_T, E^v_T, E^l_T})\) will be carried out individually to show the impact on the data in detail.

\subsection{New Incomplete Condition}

We introduce one novel incomplete condition \((\boldsymbol{E^a_T, E^v_T, E^l_T})\) utilizing the noise scheduler, which represents the addition of noise to all three modalities. This condition was not previously considered in existing research. In prior methodologies, once one modality is regarded as \textit{incomplete} or \textit{missing}, they will drop the data of this modality or set it to zero vectors, rendering the analysis of all three modalities' incompletion as both impractical and meaningless. But in real-world scenarios, this kind of incompletion is not uncommon, and the incompletion level also varies. The controllable intensity of noise allows us to report results under this extreme condition.

\subsection{Experimental Setup}
On the IEMOCAP dataset, we follow research \citep{zhao2021, zuo2023} to extract the original embeddings \(\boldsymbol{E^a}\), \(\boldsymbol{E^v}\), and \(\boldsymbol{E^l}\). The audio, visual, and lexical embeddings are 130-dim \textit{OpenSMILE} features with the configuration of \textit{IS13\_ComParE} \citep{eyben2010opensmile}, \textit{Denseface} \citep{DenseNet} embeddings extracted by a pre-trained DenseNet model of 342 dimensions, and 1024-dim \textit{BERT} \citep{devlin2018bert} word embeddings, respectively.

On the CMU-MOSEI dataset, we employ the feature extraction method from the work of Liang et al. \citep{liang2021} Audio features are 74 dimensions extracted using \textit{COVAREP} \citep{degottex2014covarep}, while visual embeddings are 35 dimensions derived from the pool5 layer of an ImageNet \citep{deng2009imagenet}-trained \textit{ResNet-152} \citep{he2016deep} model for each video frame, which underwent preprocessing steps including resizing, center cropping, and normalizing. Facial expression features are obtained using the \textit{OpenFace} \citep{eyben2010opensmile} tool. Lexical features are represented using 300-dimensional \textit{GloVe} word vectors \citep{pennington2014glove}. We adjust the balance factors $\lambda_s$ to 1, 10, and 1 for scaling the losses accordingly.

The hidden size of the LSTM structure is set to 128. The TextCNN contains 3 convolution blocks with kernel sizes of {3, 4, 5} and an output size of 128. The output size of the Invariance Encoder is also set to 128. The VAE Module includes a Transformer Encoder of 5 layers, 768 dimensions, and 16 heads as the encoder while Linear layers with dimensions of \{128, 256, 384\} as the decoder. The classifier contains three linear layers of size \{384, 128, 4\}. 
For the noise intensity, we conduct experiments on $\mathrm{{SNR}_{dB}}$ of [-10\textit{dB}, -20\textit{dB}, -30\textit{dB}, -40\textit{dB}]. 
For impulse noise, the appearance frequency $p$ is set at 0.3, with other parameters remaining unchanged.
All experiments are run on an NVIDIA A100 80GB GPU.

\begin{table*}[] 
\centering
\caption{The detailed results of our NMER model under the six incomplete conditions utilizing Gaussian noise. ``a'' means that the modality $a$ is clean, and the other two modalities ($v$ and $l$) are noise-influenced, ``Avg'' indicates the average result of six conditions.}
\renewcommand{\arraystretch}{1.5} 
\begin{tabular}{l@{\hspace{7pt}}c@{\hspace{10pt}}c c@{\hspace{10pt}}c c@{\hspace{10pt}} c c@{\hspace{10pt}} c c@{\hspace{10pt}} c c@{\hspace{10pt}} c c@{\hspace{10pt}} c c@{\hspace{10pt}}}
\toprule
\multirow{2}{*}{Dataset}   & \multirow{2}{*}{Intensity}  & \multicolumn{2}{c@{\hspace{10pt}}}{a} & \multicolumn{2}{c@{\hspace{10pt}}}{v} & \multicolumn{2}{c@{\hspace{10pt}}}{l} & \multicolumn{2}{c@{\hspace{10pt}}}{a, v} & \multicolumn{2}{c@{\hspace{10pt}}}{a, l} & \multicolumn{2}{c@{\hspace{10pt}}}{v, l} & \multicolumn{2}{c@{\hspace{10pt}}}{Avg} \\ 
& & {WA} & {UA} & {WA} & {UA} & {WA} & {UA} & {WA} & {UA} & {WA} & {UA} & {WA} & {UA} & {WA} & {UA} \\ \midrule
\multirow{4}{*}{IEMOCAP}   & -10\textit{dB} & 0.7275 & 0.7349 & 0.7440  & 0.7503 & 0.7750  & 0.7820  & 0.7432 & 0.7508 & 0.7813 & 0.7917 & 0.7877 & 0.7952 & 0.7598 & 0.7675 \\
&-20\textit{dB} & 0.6752 & 0.6873 & 0.6911 & 0.6949 & 0.7623 & 0.7735 & 0.7068 & 0.7178 & 0.7720 & 0.7846 & 0.7771 & 0.7864 & 0.7307 & 0.7406 \\
&-30\textit{dB} & 0.6574 & 0.6626 & 0.6715 & 0.6688 & 0.7513 & 0.7592 & 0.7021 & 0.7071 & 0.7649 & 0.7778 & 0.7712 & 0.7771 & 0.7197 & 0.7254 \\
&-40\textit{dB} & 0.6347 & 0.6484 & 0.6529 & 0.6511 & 0.7348 & 0.7484 & 0.6978 & 0.7024 & 0.7632 & 0.7777 & 0.7672 & 0.7773 & 0.7085 & 0.7176  \\
\midrule
\multirow{4}{*}{CMU-MOSEI }&-10\textit{dB} & 0.7274 & 0.5905 & 0.7262 & 0.6204 & 0.7787 & 0.7362 & 0.7366 & 0.6187 & 0.8030  & 0.7507 & 0.7856 & 0.7383 & 0.7596 & 0.6760  \\
&-20\textit{dB} & 0.7030  & 0.5644 & 0.6924 & 0.5885 & 0.7783 & 0.7317 & 0.7084 & 0.5965 & 0.7982 & 0.7474 & 0.7922 & 0.7395 & 0.7454 & 0.6617   \\
&-30\textit{dB} & 0.7099 & 0.5497 & 0.6959 & 0.5690  & 0.7837 & 0.7394 & 0.7094 & 0.5859 & 0.8012 & 0.7453 & 0.7924 & 0.7380  & 0.7487 & 0.6549 \\
&-40\textit{dB} & 0.7120  & 0.5579 & 0.6872 & 0.5694 & 0.7813 & 0.7263 & 0.6973 & 0.5844 & 0.8096 & 0.7456 & 0.8017 & 0.7412 & 0.7482 & 0.6543 \\ 
\bottomrule           
\end{tabular}
\label{our Gaussian condition}
\end{table*}

We utilize the AdamW \citep{adamw} as the optimizer and use the Lambda LR \citep{wu2020deep} to dynamically update the learning rate. The initial learning rate is 0.0001. The batch size is 128 and the dropout rate is 0.5.
We run experiments with 10-fold cross-validation, where each fold contains 60 epochs, and report the result on the test set. Each result is run three times and averaged to reduce the effect of random initialization of parameters. We employ the same evaluation metrics as those used in previous works \citep{zhao2021}, \citep{zuo2023}, \textit{Weighted Accuracy} (WA) \citep{WA} and \textit{Unweighted Accuracy} (UA) \citep{UA}, to assess various systems.

\subsection{Comparison and Ablation Study}
In our study, we benchmark our NMER model against four advanced MER baselines to establish its relative performance. 

1) \textbf{\textit{Modality Encoder Network (MEN)}} \citep{zhao2021}: This model serves as the \textbf{complete-modality baseline}. MEN is trained under complete modality conditions and tested on incomplete modality conditions.

2) \textbf{\textit{MCTN}} \citep{Pham2019}: MCTN uses translation-based method with cycle consistency loss to learn joint representations between every two modalities in multimodal data, which is used as a popular method.

3) \textbf{\textit{MMIN}} \citep{zhao2021}: This model employs a cascade residual AutoEncoder coupled with cycle consistency construction to learn joint representations, particularly for predicting missing modalities.

4) \textbf{\textit{IF-MMIN}} \citep{zuo2023}: An enhancement of MMIN. IF-MMIN integrates the modality invariant feature to learn robust joint representations and is recognized as a state-of-the-art incomplete modalities multimodal emotion recognition system. MCTN, MMIN, and IF-MMIN are categorized as \textbf{incomplete-modality baselines}, with training and testing both under incomplete modality conditions.

\begin{table}[ht]
\centering
\caption{The WA and UA declining on uni-modality testing.}
\renewcommand{\arraystretch}{1.5} 
\begin{tabular}{l
@{\hspace{10pt}}cc
@{\hspace{10pt}}cc
@{\hspace{10pt}}cc
@{\hspace{10pt}}}
\toprule
\multirow{2}{*}{Intensity} 
& \multicolumn{2}{c@{\hspace{10pt}}}{a} 
& \multicolumn{2}{c@{\hspace{10pt}}}{v} 
& \multicolumn{2}{c@{\hspace{10pt}}}{l} \\ 
& WA & UA & WA & UA & WA & UA \\ \midrule
 0\textit{dB}                          & 0.6693	                       & 0.6776                      &   0.5723                      & 0.5560                         & 0.6484                         & 0.6592                       \\
-10\textit{dB}       & 0.5633        & 0.5588        & 0.4252    & 0.4107      &  0.5421     & 	0.5415      \\
-20\textit{dB}                         & 0.4769                 & 0.4879                 & 	0.3941	
                & 0.3780                 & 0.4665	              & 0.4644                \\
-30\textit{dB}                         & 0.4483	                 & 0.4651                 & 0.3624
                        & 	0.3443                       & 0.3867	                   & 0.3878                       \\
-40\textit{dB}                        & 0.4161	                      & 0.4182                      &  0.3502	
                      & 0.3363                       & 0.3795	                    & 0.3333                       \\
\bottomrule   
\end{tabular}
\label{uni_modality}
\end{table}
Our experimental design not only evaluates the capability of the multimodal emotion recognition systems but also demonstrates the fine-grained control offered by our noise scheduler. We first conduct experiments on traditional six incomplete conditions, where the noise intensity progressively increases from -10\textit{dB} to -40\textit{dB}. Besides, we design the experiment on the brand new incompletion condition (\(\boldsymbol{E^a_T, E^v_T, E^l_T}\)), to explore and evaluate the potential of models under a more complex noise environment. Moreover, we launched a simple benchmark test on the IEMOCAP dataset that directly sends uni-modality data with various intensities of noise to the classifier and reports the result. This test aims to establish a benchmark for the subsequent incomplete modality multimodal emotion recognition technology, that is, how much it has improved on the basis of a single modality data and its various types of incompletion.

Additionally, ablation studies are performed to ascertain the contribution of the VAE model and the importance of the invariant features to the overall model performance. In the study, labeled as \textbf{``w/o VAE''}, we remove the VAE model. Instead, we directly concatenate the extracted emotion-specific feature \(\boldsymbol{h'}\) and the emotion-invariant feature \(\boldsymbol{H'}\), then feed this representation into the classifier. This study is designed to highlight the significant role of the VAE model in reconstructing the multimodal joint representation from noisy features. Another ablation study that removed the invariant loss, labeled as \textbf{``w/o $\mathcal{L}_{\text{inv}}$''}, aims to demonstrate the invariant features can help guide the VAE model to generate more exact representations.

\begin{table*}[]\centering
    \caption{The results of various systems utilizing Gaussian noise and adapting the new noise condition, the Impulse noise condition has a similar trend.}
\renewcommand{\arraystretch}{1.5} 
\begin{tabular}
{l@{\hspace{10pt}}l@{\hspace{10pt}}cc@{\hspace{10pt}}cc@{\hspace{10pt}}cc@{\hspace{10pt}}cc@{\hspace{10pt}}}
\toprule
\multirow{2}{*}{Dataset}   & \multirow{2}{*}{System} & \multicolumn{2}{c@{\hspace{10pt}}}{-10\textit{dB}} & \multicolumn{2}{c@{\hspace{10pt}}}{-20\textit{dB}} & \multicolumn{2}{c@{\hspace{10pt}}}{-30\textit{dB}} & \multicolumn{2}{c@{\hspace{10pt}}}{-40\textit{dB}} \\ 
                           &                         & WA             & UA       & WA             & UA       & WA               & UA     & WA              & UA      \\\midrule
\multirow{5}{*}{IEMOCAP}   & MEN                     & 0.6496         & 0.6355   & 0.5868         & 0.5545   & 0.5425	   & 0.5233      & 0.4767	    & 0.4631        \\
                           & MCTN                  & 0.6874          & 0.6930         & 0.6295               & 0.6250          & 0.5954                 & 0.6032       & 0.5469                & 0.5503        \\
                           & MMIN                    & 0.7340	    & 0.7453      & 0.6893	 & 0.7011         &0.6498 & 0.6566 & 0.6024 & 0.5997        \\
                           & IF-MMIN                & \textbf{0.7450} &\textbf{0.7546} & 0.6940 & 0.6985 & 0.6472 & 0.6531 & 0.6053 & 	0.5911         \\
                           & \textbf{Ours}                   & 0.7436 & 0.7528 & \textbf{0.6963} & \textbf{0.7026} & \textbf{0.6510} & \textbf{0.6577} & \textbf{0.6103} & \textbf{0.6136} \\        
                           \midrule
\multirow{5}{*}{CMU-MOSEI} & MEN                     & 0.7591 & 0.6471 & 0.7319 & 0.6061 & 0.7017 & 0.5707 & 0.6825 & 0.5560     \\
                           & MCTN                    & 0.7534               & 0.6635         & 0.7346               & 0.6285         & 0.7100         & 0.5844       & 0.6880                & 0.5613        \\
                           & MMIN                  &0.7566 & 0.6855 & 0.7436 & 0.6459 & 0.7301 & 0.6081 & 0.7204 & 0.5625        \\
                           & IF-MMIN                &\textbf{0.7680}  & \textbf{0.6893} & \textbf{0.7588} & 0.6467 & 0.7327 & 0.6170 & 0.7240 & 0.5706         \\
                           & \textbf{Ours}                    & 0.7651 	          &  0.6842	        &  0.7570   & \textbf{0.6513}         & \textbf{0.7341}       & \textbf{0.6181}       & \textbf{0.7266}        & \textbf{0.5735}       \\ \bottomrule
\end{tabular}
\label{declining}
\end{table*}

\subsection{Results}
\label{results}
\subsubsection{Results About Models} \

The average results of the comparative studies, across six incomplete conditions, four noise intensities and two noise types, are shown in Table~\ref{comp_1}, including \{a\}, \{v\}, \{l\}, \{a, v\}, \{a, l\}, \{v, l\},\footnote{\{$\cdot$\} means the clean modality. The noisy modalities are omitted.}, with Gaussian and impulse noise at four levels of intensities.

1) The NMER model outperforms most baseline models across various noise intensities and types. For instance, under -10\textit{dB} Gaussian noise in the IEMOCAP dataset, NMER's WA is 0.7598 compared to 0.7120 for MEN, 0.7215 for MCTN, 0.7551 for MMIN, and 0.7543 for IF-MMIN. UA results show a similar pattern. 

2) There is a notable difference in accuracy (mainly in high-level noise conditions, for example -40\textit{dB}, about 6\% - 10\%) between the complete-modality baseline (MEN) and the incomplete-modality baselines (MCTN, MMIN, and IF-MMIN) as shown in Table \ref{comp_1}, which indicates the importance of using incomplete training data. 

3) The results of our model on both datasets under six incomplete conditions are detailed in Table~\ref{our Gaussian condition} (taking Gaussian noise as an example), the WA values for single clean modality conditions \{$a$\} and \{$v$\} drop by about 9\% with the noise intensity increases from -10\textit{dB} to -40\textit{dB}, whereas in two clean modalities conditions \{$a$, $v$\}, \{$v$, $l$\}, and \{$v$, $l$\}, the decrease is slower, ranging from about 2\% - 4\%. When modality $a$ or $v$ is noise-influenced, the performance decreases are roughly equivalent.
This indicates that the resistance of the model to noise worsens significantly when only one modality is noise-free. 
However, the model shows relative robustness when the two modalities are noise-free. Notably, modality $l$ plays a critical role in maintaining performance even when other modalities are compromised by noise. The WA value decreased about 4\% when only modality \textit{l} is clean, and about 2\% when two modalities including $l$ are noise-free (condition \{$v$, $l$\} and \{$v$, $l$\}).

4) The WA and UA results on the CMU-MOSEI dataset, as shown in Table~\ref{comp_1}, exhibit different trends and absolute values compared to those on the IEMOCAP dataset, primarily due to the imbalanced label distribution  (as shown in Table~\ref{label distribution}). In such scenarios, classifiers are inclined to categorize samples into the more samples class because this strategy statistically enhances their accuracy performance.  Because of the way the WA value is calculated, if the classification performance of the larger number samples class is good, then even if the noise increases, the weighted accuracy (WA) may not decrease significantly as long as the prediction for that class remains accurate. Because that class contributes more to the overall accuracy, the decline in the accuracy of fewer sample class is offset. These reasons result in the different variation trends of WA and UA values.

\subsubsection{Results About the Noise Scheduler}\

1) It is evident that the performance of all models deteriorates as noise intensity increases, which aligns with the expectation that increasing noise levels obscure the original data's information. For instance, in Table~\ref{comp_1}, the WA of the IF-MMIN system on the IEMOCAP dataset under Gaussian noise declines from 0.7543 at -10\textit{dB} to 0.7048 at -40\textit{dB}. These results highlight the effectiveness of our noise scheduler, which allows for precise control over noise intensities and the simulation of various noise conditions.

2) From the accuracy declining and the noise intensity increasing process in Table~\ref{comp_1}, it becomes crucial to recognize the shortcomings of previous studies that used zero vectors or drops data to simulate noise. These methods lack rationality and precise control over noise intensity and fail to enable meaningful comparisons across different noise types and intensities. In contrast, the experiments demonstrate the utility of our method in managing noises, achieving more comprehensive evaluations of the models.

3) In Table~\ref{uni_modality}, we test the uni-modality result on the IEMOCAP dataset using Gaussian noise. The results in the first row show that different modalities have different quantities of information and difficulties of recognition. The performance of different modalities at the same noise level shows that different data types (audio, video, lexical) have different sensitivity to ambient noise. For example, the video ($v$) modality has a greater performance degradation at -10\textit{dB} (about 15\%) than the audio ($a$) and lexical ($l$) modalities (about 10\%).
With the incremental noise intensities (from 0\textit{dB} to -40\textit{dB}), the accuracy of each modality has its own downtrend. This downward trend highlights the difficulty of effective emotion recognition in high-noise environments, while also providing data support for the importance of multimodal fusion strategies.
This table emphasizes the necessity and superiority of multimodal joint representation learning at the same time. When one mode is seriously disturbed, other modes may still be able to maintain good recognition performance, thus improving the overall emotion recognition accuracy.

\subsubsection{Results About New Incomplete Condition}\

The result of the new incomplete condition is listed in Table~\ref{declining}.

1) The introduction of scenarios with full-modality noise interference in multimodal emotion recognition presents a new challenge, reflecting complex real-world environments where multiple information sources can simultaneously experience quality degradation. This expands the boundaries of current research. Traditionally, models could rely on at least one clean modality. However, this new setup eliminates such possibilities, demanding greater robustness and denoising capability from the models themselves. With noise affecting all modalities, the accuracy significantly decreases. For instance, in Table~\ref{declining}, on the IEMOCAP dataset, the WA value of the MMIN model falls by about 4\% from -10\textit{dB} to -20\textit{dB} noise levels and further declines by approximately 13\% at -40\textit{dB}. It indicates that the traditional multimodal fusion strategy may not be able to effectively deal with the simultaneous modal degradation, exposing the limitations of existing techniques against cross-modal interference.

2) There is a noticeable acceleration in performance degradation as noise intensity increases, which may be caused by the non-linear nature of emotion data and the complex response of models to varying noise levels. This information loss complicates the interaction between different modalities. Notably, all models demonstrate a similar trend under this condition. Future research should thus concentrate on studying how noise variations affect data and exploring more strategies to mitigate the impact of diverse noise.

\subsubsection{Ablation Results}\

The result of the last two rows in Table \ref{comp_1} shows the crucial role of the VAE model and the invariant features' guidance.
For example, in the IEMOCAP dataset, under Gaussian noise, the WA of two ablation studies achieves 0.7382 and 0.7401 on the noise intensity -10\textit{dB} whereas the NMER achieves 0.7598. The outcomes clearly indicate the inclusion of the contributions that these parts provided to the overall performance of our NMER model.

\section{Conclusion}
\label{sec:con}
This work proposed a Noise-robust Multimodal Emotion Recognition model (NMER) that effectively mitigates the impact of incomplete data and reconstructs the robust multimodal joint representations from incomplete data.
Experimental results show that Our noise scheduler can effectively create different types and intensities of noisy data to simulate various noise corruptions. Our NMER model achieves robust performance across various incomplete situations.  
Notably, there exist huge challenges in the new incomplete condition, which proposes more requirements to the MER systems.  Our research is only the first step in addressing this problem.

\section{Acknowledgements}
The research by Rui Liu was funded by the Young Scientists Fund of the National Natural Science Foundation of China (No. 62206136), the Guangdong Provincial Key Laboratory of Human Digital Twin (No. 2022B1212010004), and the ``Inner Mongolia Science and Technology Achievement Transfer and Transformation Demonstration Zone, University Collaborative Innovation Base, and University Entrepreneurship Training Base'' construction Project (Supercomputing Power Project) (21300-231510). The work is also supported by the following fundings: National Natural Science Foundation of China (62066033), the Outstanding Youth Foundation of the Natural Science Foundation of Inner Mongolia (2022JQ05), the Science and Technology Program of Inner Mongolia Autonomous Region (2021GG0158), the Hohhot Collaborative Innovation Project for Universities and Institutes, the Youth Science and Technology Talent Cultivation Project of Inner Mongolia University (21221505), the fund of Supporting the Reform and Development of Local Universities (Disciplinary Construction) and the special research project of First-class Discipline of Inner Mongolia A. R. of China under Grant (YLXKZX-ND-036).
College of Computer Science, Inner Mongolia University, National \& Local Joint Engineering Research Center of Intelligent Information Processing Technology for Mongolian, and Inner Mongolia Key Laboratory of Multilingual Artificial Intelligence Technology provided equipment support to this work.


\bibliographystyle{ACM-Reference-Format}
\balance
\bibliography{sample-base}









\end{document}